\documentclass{comjnl}

\usepackage{comment}
\usepackage{fancyvrb}
\usepackage{graphicx}
\usepackage{hyperref}
\usepackage[section]{placeins}
\usepackage{gensymb}

\begin{document}

\title[A Roadmap for Developing the ``SP Machine'']{A Roadmap for the Development of the ``SP Machine'' for Artificial Intelligence}

\author{Vasile Palade}
\affiliation{School of Computing, Electronics and Mathematics, Coventry University, UK}

\author{J. Gerard Wolff}
\affiliation{CognitionResearch.org, Menai Bridge, UK. Phone: 01248 712962}

\email{Corresponding author: jgw@cognitionresearch.org}

\shortauthors{V. Palade and J. G. Wolff}

\received{00 Month 2018}
\revised{00 Month 2018}


\begin{abstract}

This paper describes a roadmap for the development of the {\em SP Machine}, based on the {\em SP Theory of Intelligence} and its realisation in the {\em SP Computer Model}. The SP Machine will be developed initially as a software virtual machine with high levels of parallel processing, hosted on a high-performance computer. The system should help users visualise knowledge structures and processing. Research is needed into how the system may discover low-level features in speech and in images. Strengths of the SP System in the processing of natural language may be augmented, in conjunction with the further development of the SP System's strengths in unsupervised learning. Strengths of the SP System in pattern recognition may be developed for computer vision. Work is needed on the representation of numbers and the performance of arithmetic processes. A computer model is needed of {\em SP-Neural}, the version of the SP Theory expressed in terms of neurons and their inter-connections. The SP Machine has potential in many areas of application, several of which may be realised on short-to-medium timescales.

\end{abstract}

\maketitle

\section{Introduction}

The main purpose of this paper is to describe a roadmap for the development of an industrial-strength {\em SP Machine}, a `new generation' computing system with intelligence based on the {\em SP Theory of Intelligence} and its realisation in the {\em SP Computer Model}, two things which, together, may be referred to as the {\em SP System}. That system, outlined briefly in Section \ref{sp_theory_and_model_section} and more fully in \ref{outline_of_sp_system_appendix} is the product of a long-term programme of research by the second author, begun in 1987 with a break between early 2006 and late 2012.

This paper is radically different from papers that report the procedures and results of one or more experiments that have been conducted already, or papers that review a range of such reports in a given field. Instead, this paper provides a roadmap for future developments, looking forward to a whole programme of research to develop the SP Machine.

Early ideas about the SP System were described in this journal in \cite{wolff_1990}. Relatively full descriptions of the SP System as it is now may be found in the book {\em Unifying Computing and Cognition} \cite{wolff_2006} and the paper ``The SP Theory of intelligence: an overview'' \cite{sp_extended_overview}. Details of several other papers, with download links, may be found on \href{http://www.cognitionresearch.org/sp.htm}{www.cognitionresearch.org/sp.htm}.

At all but the earliest stages, development and testing of the SP Computer Model has been a valuable aid to theorising: helping to reduce vagueness in the theory, as a means of testing whether ideas work as anticipated, and as a means of demonstrating what the system can do. It is envisaged that, as the SP Machine is developed, there will be similar benefits yielding further insights and further development of the SP Theory.

It is anticipated that, when it is mature, there will be many applications for the SP Machine. Some of these potential applications are described in \ref{potential_benefits_applications_appendix}.

\subsection{The SP Theory and Computer Model}\label{sp_theory_and_model_section}

{\em The SP Theory of Intelligence, with its realisation in the SP Computer Model, is a unique attempt to simplify and integrate observations and concepts across artificial intelligence, mainstream computing, mathematics, and human learning, perception and cognition, with information compression as a unifying theme.} It is described in \ref{outline_of_sp_system_appendix} with enough detail to allow the rest of the paper to be understood.

The SP System has several distinctive features and advantages compared with AI-related alternatives \cite{sp_alternatives}.

The most distinctive feature of the SP System is the powerful concept of {\em SP-multiple-alignment}, borrowed and adapted from the concept of `multiple sequence alignment' in bioinformatics. A key advantage of the SP-multiple-alignment framework as it has been developed in the SP System is its versatility in aspects of intelligence, its versatility in the representation of diverse kinds of knowledge, and its potential for the seamless integration of diverse aspects of intelligence and diverse kinds of knowledge, in any combination. Seamless integration of diverse aspects of intelligence and diverse kinds of knowledge appears to be {\em essential} in any artificial system that aspires to the fluidity, versatility, and adaptability of human intelligence.

\subsection{The SP Machine}

\begin{figure*}[!htbp]
\centering
\includegraphics[width=0.9\textwidth]{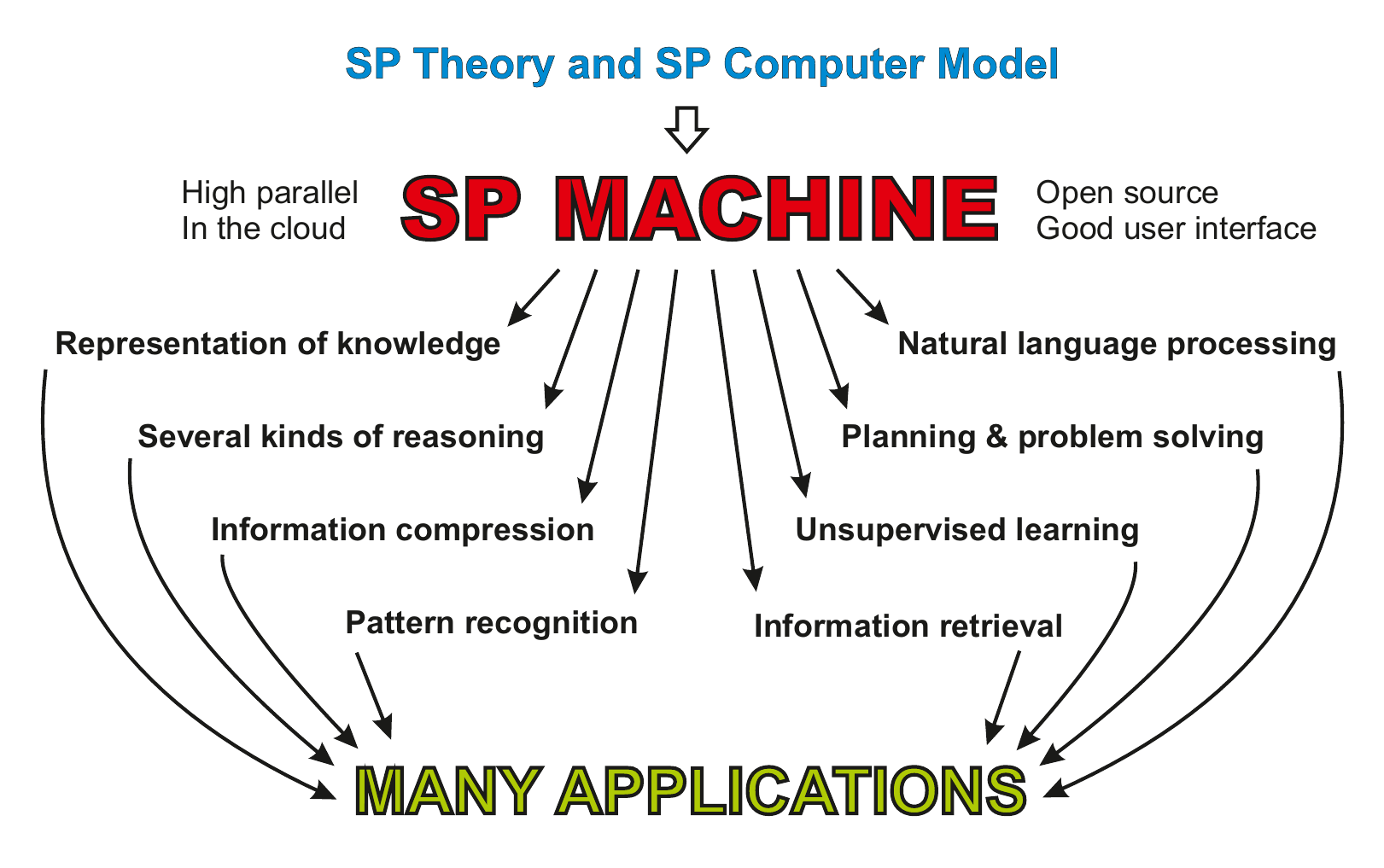}
\caption{A schematic view of how the SP Machine may develop from the SP Theory and the SP Computer Model. Adapted, with permission, from Figure 6 in \protect\cite{sp_big_data}.}
\label{sp_machine_development_figure}
\end{figure*}

It is anticipated that a fully-developed SP Machine will provide a means of realising the several potential benefits and applications of the SP System, as shown schematically in Figure \ref{sp_machine_development_figure}. Some of these potential benefits and applications are described in \cite{sp_benefits_apps} and in the several papers referenced in \ref{potential_benefits_applications_appendix}.

It is envisaged that, initially, the SP Machine will be developed as a highly-parallel software virtual machine, hosted on an existing high-performance computer, and with a user interface that helps users to visualise knowledge structures in the system and how they are processed.

Although the SP Computer Model is still useful as a vehicle for research (Section \ref{research_with_sp_computer_model_section}), the main function of the SP Machine as just described would be to open things up to facilitate research across a wider range of issues. It may also prove useful as a means of demonstrating what the SP System can do.

At some stage, there is likely to be a case for the development of new hardware, dedicated to structures and processes in the SP System, and with potential for substantial gains in efficiency and performance (\cite[Section IX]{sp_big_data}, \cite[Section III]{sp_autonomous_robots}).

\subsection{Presentation}

After a summary of the research strategy which is envisaged (Section \ref{overview_of_research_strategy_section}), several sections describe things that need to be done in the development of the SP Machine, including a description of problems to be solved, and potential solutions to those problems, including some areas of application that may yield benefits on relatively short timescales.

Four of the sections are about pieces of ``unfinished business'' in the development of the SP Machine, outlined in \cite[Section 3.3]{sp_extended_overview}: processing information in two or more dimensions (Section \ref{processing_2d+_information_section}); recognition of perceptual features in speech and visual images (Section \ref{low_level_features_in_speech_and_vision_section}); unsupervised learning (Section \ref{unsupervised_learning_development_section}); and processing numbers (Section \ref{processing_numbers_section}).

\subsection{An Invitation}

In accordance with the saying that ``A camel is a horse designed by a committee'', the development of a scientific theory, like writing a novel or composing a symphony, is a task that is not easily shared. But now that the SP Theory is relatively mature, it should become easier progressively for other researchers to work on the several avenues that now need to be explored, without treading on each others' toes.

The involvement of other researchers would be very much welcomed. The authors of this paper would be happy to discuss possibilities, and the necessary facilities.

\section{Background}\label{background_section}

The {\em SP Theory of Intelligence} and its realisation in the {\em SP Computer Model} is the product of an extended programme of research by the second author.

In accordance with widely-accepted principles in science, the overarching goal of this programme of research has been the simplification and integration of observations and concepts across artificial intelligence, mainstream computing, mathematics, and human learning, perception, and cognition, with information compression as a unifying theme.

In this respect, the research contrasts sharply with the widespread and longstanding tendency for research in AI to become fragmented into many subfields, described very well by Pamela McCorduck \cite{mccorduck_2004}.\footnote{``The goals once articulated with debonair intellectual verve by AI pioneers appeared unreachable ...~Subfields broke off---vision, robotics, natural language processing, machine learning, decision theory---to pursue singular goals in solitary splendor, without reference to other kinds of intelligent behaviour.'' \cite[p.~417]{mccorduck_2004}.} The research is much more in the spirit of exhortations---by Allen Newell \cite{newell_1973,newell_1990}, by researchers in `artificial general intelligence' (e.g., \cite{everitt_etal_2017}), \cite{langley_2012}, and others)---for researchers in AI and related areas to adopt a high-level global view of their field, and to seek unifying principles across wide areas.

The SP Theory draws on an earlier programme of research (summarised in \cite{wolff_1988}) developing computer models of how a child learns his or her first language or languages. The emphasis on hierarchical structures in those models, excluding other kinds of structure, proved unsuitable for the SP programme of research because of the much wider scope of the later research. But information compression remains a key principle in the SP research as it proved to be in the earlier research.

After several years of work by the second author---reading, thinking, and experimentation with computer models---it gradually became clear that a modified version of `multiple sequence alignment' from bioinformatics had the potential to provide a good unifying framework across artificial intelligence, mainstream computing, mathematics, and human learning, perception, and cognition, with information compression as a unifying theme.

This new framework---{\em SP-multiple-alignment}---has been progressively refined to provide a robust computational framework for diverse aspects of intelligence, for the representation of diverse kinds of knowledge, and for their seamless integration in any combination, where ``any combination'' means a set comprising all the afore-mentioned aspects of intelligence and diverse kinds of knowledge, or any subset of that set.

\section{Overview of Research Strategy}\label{overview_of_research_strategy_section}

As already mentioned, the `SP' research, up to now, is a unique attempt to simplify and integrate observations and concepts across a broad canvass, with information compression as a unifying theme. We envisage that the same principles will apply in the future developments described in this paper.

Already, this long-term objective is showing many potential dividends:

\begin{itemize}

    \item The strengths of SP-multiple-alignment (\cite[Chapter 3]{wolff_2006}, \cite[Section 4]{sp_extended_overview}) in the simplification and integration of observations and concepts across a broad canvass is a major step forward with clear potential in the long-sought-after quest for general, human-level artificial intelligence, and for developing an understanding of human intelligence.

    \item As described in \ref{potential_benefits_applications_appendix}, the SP System has many potential benefits and applications.

\end{itemize}

\subsection{Maintain a Strong Focus on Simplicity and Power}\label{strong_focus_section}

The long view of this research contrasts with, for example, most research on deep learning in artificial neural networks where undoubted successes obscure several weaknesses in the framework (\ref{sp_distinctive_features_advantages_appendix}), suggesting that deep learning is likely to be a blind alley in any search for general human-level artificial intelligence.

For these kinds of reasons, researchers working with the SP concepts are encouraged to maintain a strong focus on maintaining a favourable combination of conceptual simplicity with descriptive or explanatory power, resisting the temptation to take short cuts or to introduce ideas that do not integrate well with the SP framework. That said, the following kinds of development, which are not mutually exclusive, appear to be reasonable:

\begin{itemize}

    \item There may be opportunities to develop SP capabilities with industrial strength for practical applications, without corrupting the core concepts (Section \ref{apps_on_short_timescales_section}). An example is the potential of the SP Machine to serve as a database management system with versatility and intelligence \cite{wolff_sp_intelligent_database}.

    \item There may be a case, on occasion, to develop a hybrid system as a stop-gap to facilitate practical applications, pending the development of a solution that is theoretically clean. An example in computer vision would be the use of existing procedures for the detection of low-level features in images (such as lines and corners) pending an analysis in terms of SP concepts (Section \ref{low_level_features_in_speech_and_vision_section}).

\end{itemize}

\subsection{Scheduling}

As described later, some things are probably best done before others. These `ground-clearing' tasks are described early in what follows. When those tasks have been completed, it should become easier to develop two or more aspects of the SP Machine in parallel, in tasks described in later subsections.

When the SP System has been brought close to the point where it may prosper in a commercial environment, perhaps via applications that may be developed on relatively short timescales (Section \ref{apps_on_short_timescales_section}), the system may be refined in specific application areas.

\section{The Creation of a Highly-Parallel SP Machine}\label{high_parallel_sp_machine_section}

The proposed development of the SP Machine should begin with SP71, the latest version of the SP Computer Model, outlined in \ref{sp_computer_model_appendix}. Instructions for obtaining the source code are in \ref{source_code_for_SP71_appendix}.

To get things going, the first stage, as outlined in this main section, is probably the development of the SP Machine with high levels of parallel processing, and then, in Sections \ref{user_friendly_user_interface_section} and \ref{visualisation_of_structures_and_processes_section}, the creation of a user interface that is easy to use and that facilitates the visualisation of knowledge structures and the visualisation of the processing of knowledge.

\subsection{Is Parallel Processing Really Needed?}\label{is_parallel_processing_needed_section}

A substantial advantage of the SP System, compared with, for example, deep learning in artificial neural networks \cite{schmidhuber_2015}, is that, on the strength of evidence to date, the SP Machine is likely to be very much less demanding of computational resources than deep learning systems \cite[Section V-E]{sp_alternatives}. And there appears to be potential to speed up processing by the judicious use of indexing, as outlined in Section \ref{indexing_for_speed_section}.

More generally, there appears to be potential in the SP System for very substantial gains in computational efficiency compared with conventional computers: by the use of a technique called ``model-based coding'' for the efficient transmission of data \cite[Section VIII]{sp_big_data}; and by exploiting the statistical information which is gathered as a by-product of how the system works \cite[Section IX]{sp_big_data}.

So it is reasonable to ask whether the existing computer model (SP71), or some successor to it, might be speedy enough as it is, without the need for parallel processing? Here are some answers to that question which are not mutually exclusive:

\begin{itemize}

    \item Although, as mentioned above, the SP Machine has the potential to be relatively frugal in its computational demands, its full potential in that respect will probably not be fully realised until new hardware has been developed to take full advantage of opportunities in the structure and workings of the SP System.

    \item The afore-mentioned gains in efficiency in the transmission of information via model-based coding, will not be fully realised until unsupervised learning in the SP System is more fully developed.

    \item Although, with AI problems, the SP System is likely to be much more efficient than deep learning systems, AI problems are normally quite computationally demanding compared with, for example, school-level mathematics. For that kind of reason, parallel processing is likely to be needed to achieve acceptable run times when the SP Machine is applied to what is normally the relatively demanding task of learning in unsupervised mode, and perhaps also to such things as pattern recognition and the processing of natural languages.

    \item When the SP Machine has been generalised to work with two-dimensional patterns, for reasons described in Section \ref{introduction_of_2d_patterns_section}, it is likely that parallel processing will be needed to speed up processing in what is likely to be the computationally-intensive process of finding good full or partial matches between such patterns.

    \item Parallel processing will almost certainly be needed if the SP Machine is to become an industrial-strength database management system with intelligence, as described in \cite{wolff_sp_intelligent_database}.

\end{itemize}

In short, there are several reasons why the SP Machine should exploit parallel processing.

\subsection{How Parallel Processing May Be Applied}\label{how_parallel_processing_may_be_applied_section}

The effective application of parallel processing will require careful analysis of the SP71 computer model (\ref{sp_computer_model_appendix}) to identify parts of the program where parallel processing may be applied.

A consistency model, such as Leslie Lamport's ``sequential consistency model'' \cite{lamport_1979}, or one of its successors, will be needed to ensure that any parallel version of the SP Computer Model produces the same results as the original model.

There may be a case for using an appropriate calculus or algebra. such as ``Communicating Sequential Processes'' \cite{abdallah_etal_2005}, to facilitate reasoning about the workings of the parallel system.

At this stage, it seems likely that parallel processing may be applied most effectively via a MapReduce technique (see, for example, \cite{dean_ghemawat_2008}). It may perhaps be applied in these parts of the program:

\begin{itemize}

    \item With {\em the building of SP-multiple-alignments} (\ref{building_multiple_alignments_appendix}), a recurring operation is searching for good full or partial matches between one ``driving'' SP-pattern and each of a set of ``target'' SP-patterns (\ref{full_and_partial_matches_appendix}), where the set of target SP-patterns may become very large.

        Here, the ``map'' phase would be applying the search in parallel to each pairing of a driving SP-pattern with a target SP-pattern, while the ``reduce'' phase would be identifying the most successful matches and selecting a subset for further processing.

    \item With {\em unsupervised learning}, each SP-pattern in a relatively large set of New SP-patterns is processed via a search for SP-multiple-alignments that allow the given New SP-pattern to be encoded economically.

        Here, the ``map'' phase would be the application of the search process to all the New SP-patterns in parallel, and the ``reduce'' phase would be a process of sifting and sorting to create one or more alternative grammars for the set of New SP-patterns.

\end{itemize}

There may be other opportunities for applying the MapReduce technique in the SP Machine. There may also be opportunities for the use of other techniques such as pipelining (see, for example, \cite{lam_1988}).

\subsection{Computing Environments for the Application of Parallel Processing}

For this development, we have identified two main kinds of computing environment that may be used for the application of parallel processing:

\begin{itemize}

    \item {\em A stand-alone computer cluster}. A server with one or more multi-core chips and appropriate software is a promising vehicle for the research. We believe this is likely to be better than most large-scale `supercomputing' facilities, because it would give researchers unfettered access to develop programs and to test them frequently, without restrictions.

    \item {\em Highly-parallel search mechanisms in a search engine}. An interesting alternative would be to replace the low-level search mechanisms in the SP Machine with highly-parallel search mechanisms in an existing search engine. This would supercharge the SP Machine with the power of the given search engine, and would in effect add intelligence to the search engine. Here is a little more detail:

        \begin{enumerate}

            \item It appears that the most computationally-intensive part of the SP Computer Model is the process of finding good full and partial matches between pairs of patterns, as described in \cite[Appendix A]{wolff_2006}.

            \item Finding good full and partial matches between pairs of patterns is one of the things that can be achieved by any of the leading search engines.

            \item With any of the leading search engines, that kind of searching can be done with remarkable speed.

            \item Part of this speed is due to indexing, but the application of high levels of parallel processing is at least as important.

            \item These observations suggest that it makes sense to try to speed up the SP System via the application of highly-parallel search mechanisms in a search engine, including the use of indexing (Section \ref{indexing_for_speed_section}).

        \end{enumerate}

\end{itemize}

The SP Machine, hosted on a search engine, may approach human-like intelligence with access to all the knowledge that has been indexed in the given search engine. Looking further ahead, all search engines may be adapted in this way, and some of the processing may be offloaded to PCs and other computers around the world, in the manner of volunteer computing projects such as SETI@home (\href{https://bit.ly/2o37khu}{bit.ly/2o37khu}) and Folding@home (\href{https://bit.ly/2AJmXTZ}{bit.ly/2AJmXTZ}).

Many benefits may flow from artificial intelligence like that, in the internet and beyond. But there are also potential dangers and other problems which will need careful scrutiny and much debate.

\section{Indexing and Hash Coding as Possible Means of Speeding Up Processing}\label{indexing_for_speed_section}

A prominent feature of the SP System, both in the way it does unsupervised learning and the way it builds SP-multiple-alignments, is a process of searching for SP-symbols that match each other, which is normally part of a process of searching for full or partial matches between SP-patterns.

Since that kind of searching for matches between symbols is often done repeatedly with the same sets of New and Old SP-patterns, there is potential for substantial gains in computational efficiency by recording matches as they are found and using that information to by-pass searching in later processing. The record of matches that have been found is, in effect, an index into the data. This may be done with matches between individual SP-symbols, and may perhaps also be done for matches between SP-patterns.

In the process of finding the entry, if any, for any given symbol in the index mentioned above, there may be a case for introducing hash coding as a means of increasing the speed of that processing, as described in, for example, \cite[pp.~61-65]{thesen_1978}.

As outlined in Section \ref{implementation_of_sp_neural_section}, similar ideas may be applied in {\em SP-Neural}, a version of the SP Theory expressed in terms of neurons and their interconnections.

\section{Research That May Be Done With the SP Computer Model}\label{research_with_sp_computer_model_section}

It is envisaged that the highly-parallel SP Machine with a `friendly' user interface will be a good foundation for further development of the SP Machine. But pending that development, quite a lot can be done with the SP Computer Model. The main areas that may need to wait for completion of the highly-parallel SP Machine are:

\begin{itemize}

    \item {\em Processing information in two or more dimensions}, Section \ref{processing_2d+_information_section}. This is likely to be relatively demanding for computing power.

    \item {\em The processing of natural language}, Section \ref{processing_natural_language_section}. Progress can be made in experimenting with small-scale samples of English-like artificial languages that exhibit structures of interest. But it is likely that significantly more power will be needed for the processing of realistically-large bodies of natural language.

    \item {\em Unsupervised learning}, Section \ref{unsupervised_learning_development_section}. As with natural language, some progress can be made with small examples. But unsupervised learning with realistically large bodies of data is likely to require the power of the highly-parallel SP Machine.

    \item {\em Computer vision}, Section \ref{computer_vision_section}. As with the processing of 2D SP-patterns, computer vision is likely to be relatively demanding of computer power.

    \item {\em SP-Neural}, Section \ref{sp-neural_section}. Early work in the development of a computer model of SP-Neural may be done with the SP Computer Model. But more power will probably be needed later.

    \item {\em Development of applications of the SP Machine}, Section \ref{devt_apps_sp_machine_section}. It will be best to develop most of the SP potential applications on the highly-parallel SP Machine.

\end{itemize}

\section{Development of a `Friendly' User Interface}\label{user_friendly_user_interface_section}

The user interface for the SP71 computer model is the product of {\em ad hoc} developments in response to evolving needs, without any attempt at a coherent and streamlined design.

This has not mattered for the one user (the second author) who originated each feature and has learned the workings of the user interface as it evolved. However, as the SP System is opened up to new users, a more intuitive and `friendly' user interface is needed, taking advantage of advanced graphical techniques that are available now.

Since all users, both local and remote, may access the SP Machine via a web browser, and to eliminate the need for users to download any software, it is envisaged that the new user interface will be developed using HTML5 or one of its successors.

\section{Development to Facilitate Visualisation of Structures and Processes}\label{visualisation_of_structures_and_processes_section}

An important concern with artificial intelligence is that it should be possible to inspect and understand how an AI system stores its knowledge and how it reaches its conclusions. This is important where an AI system may have an impact on the safety of people or where its performance and conclusions may have a bearing on court cases.

A major strength of the SP System is that all its knowledge is open to inspection and it provides an audit trail for all its processing \cite[Section XI]{sp_big_data}.

Although this information is provided by the system, there is probably scope for improving its presentation, so that structures and processes can be seen easily and so that users can find easily what they are looking for.

\section{Processing Information in Two or More Dimensions}\label{processing_2d+_information_section}

It has been recognised for some time that the SP Machine should be generalised to work with information in two dimensions and probably more \cite[Section 13.2.1]{wolff_2006}. How this may be done is discussed in the following subsections.

\subsection{Introduction of Two-Dimensional SP-Patterns}\label{introduction_of_2d_patterns_section}

The SP Computer Model as it is now works only with one-dimensional patterns. This is good for modelling such things as natural language texts, grammars for natural or artificial languages, or logical or mathematical expressions, but it is not well-suited to the representation and processing of such things as photographs, paintings, drawings, and diagrams.

Those last-mentioned kinds of entity provide the main motive for introducing 2D SP-patterns into the SP System. But 2D SP-patterns may also serve in the definition and execution of procedures that operate in parallel \cite[Sections V-G, V-H, and V-I, and Appendix C]{sp_autonomous_robots}, and, as described in Section \ref{modelling_3d_structures_section}, 2D SP-patterns can play a useful part in the representation and processing of 3D structures.

It is a simple enough matter to add 2D SP-patterns to the SP System. What is more challenging is how the concept of SP-multiple-alignment may be generalised to incorporate 2D SP-patterns as well as 1D SP-patterns, how to represent SP-multiple-alignments which include 2D SP-patterns on a 2D computer screen or a printed page, and how unsupervised learning may work with 2D SP-patterns as well as 1-D SP-patterns.

Some clues may be gleaned from how, in digital photography, pictures that overlap each other may be stitched together to create a panorama. However, it seems unlikely that software that is designed for the stitching together of overlapping pictures will have the flexibility needed to achieve the 2D equivalent of the processes in SP71 that can find good full and partial matches between two 1D SP-patterns, including cases where there are substantial differences between the SP-patterns (\ref{full_and_partial_matches_appendix}).

Generalising that process is likely to be a significant challenge. It is likely also that any reasonably good solution will absorb quite a lot of computing power. To achieve useful speeds in processing, it is likely that high levels of parallel processing will be required, as noted in Section \ref{is_parallel_processing_needed_section}.

\subsection{Modelling Structures in Three Dimensions}\label{modelling_3d_structures_section}

At first sight, the obvious way to model 3D structures in the SP System is to add 3D SP-patterns to the system. But that is probably a step too far, partly because of the complexity of matching patterns in three-dimensions and the difficulty of representing SP-multiple-alignments of 3D SP-patterns, and partly for reasons described here:

\begin{quote}

    ``Patterns in one or two dimensions may serve well enough to represent concepts with three or more dimensions. There is a long tradition of representing buildings and other engineering artifacts with two-dimensional plans and elevations.'' \cite[Section 13.2.2]{wolff_2006}.

\end{quote}

\noindent and

\begin{quote}

    ``...~there is a distinct tendency for [the brain] to be organised into layers of cells such as those found in the cortex and in the lateral geniculate body. Although these are often folded into three dimensions, they are topologically equivalent to two-dimensional sheets.~...~these sheets of neurons would provide a natural home for SP-patterns.'' {{\em ibid}.}.

\end{quote}

\sloppy The case for using 2D SP-patterns as a means of building and representing 3D structures is also strengthened by the existence of applications that work like that, such as ``Big Object Base'' \href{http://bit.ly/1gwuIfa}{bit.ly/1gwuIfa}, ``Camera 3D'' \href{http://bit.ly/1iSEqZu}{bit.ly/1iSEqZu}, and ``PhotoModeler'' \href{http://bit.ly/MDj70X}{bit.ly/MDj70X}, and also Google's ``Streetview'',\footnote{See ``Google Street View'' in ``Google Maps'', 2018-05-11.} which builds 3D models of street plans from 2D photographs. There is relevant discussion in \cite[Sections 6.1 and 6.2]{sp_vision}.

\subsection{Modelling Structures in Four or More Dimensions}\label{four_or_more_dimensions_section}

Although there are four dimensions in our everyday experience---three dimensions of space and one of time---it seems likely that, normally, people don't attempt to integrate them in the manner of Einstein's space-time but concentrate on a subset of the four dimensions, such as the succession of 2D images on our retinas as we walk through a building or drive around a town. It seems likely that, for most applications, that kind of partial integration would work well enough.

It seems likely that an important application of the SP Machine will be in learning 2D and 3D structures that change across the time dimension---from video recordings in which there will be a succession of 2D frames. With 3D videos it seems likely that preprocessing will be needed to extract structural information from each stereo pair of frames, followed by processing of the time dimension to extract regularities across time.

Extracting regularities across time is something that is likely to play an important part in how the SP System comes to isolate such things as people, birds, cars, and so on as discrete entities, and classes of such things, in the system's conceptual world, and it is likely to be important in encoding such concepts as motion and speed \cite[Section 5]{sp_vision}.

If for any reason there is a need to represent structures in five or more dimensions, it is probably best to borrow techniques used by mathematicians for dealing with such structures.

\subsection{Unordered Patterns or Sets}

As noted in \ref{patterns_symbols_and_mas_appendix}, there may be a case for introducing unordered SP-patterns or sets, where the order of the SP-symbols is not significant. But that may not be necessary since sets of New SP-patterns have no intrinsic order and that may be sufficient for the modelling of unordered groupings, as described in \cite[Section 3.4]{wolff_medical_diagnosis}.

\section{Recognition of Low-Level Perceptual Features in Speech and Images}\label{low_level_features_in_speech_and_vision_section}

As noted in \ref{patterns_symbols_and_mas_appendix}, the SP System is designed to represent and process stored knowledge and sensory data expressed as atomic {\em SP-symbols} in {\em SP-patterns}. So at first sight, the SP System should be able to work with any digital information, meaning information that is founded on the atomic symbols `\texttt{0}' and `\texttt{1}'. But there are two main complications:

\begin{itemize}

    \item Much digital information drawn relatively directly from the environment is encoded as, for example, hexadecimal numbers representing such things as levels of brightness in an image. Until the SP System can deal effectively with numbers (Section \ref{processing_numbers_section}), it will fail to interpret such information in an appropriate manner.

    \item In principle, all such encoded information may be translated into a simpler kind of representation in which, for example, the brightness in any given portion of an image would be represented by the density of `\texttt{0}'s or `\texttt{1}'s, much as in old-style black and white newspaper photograph \cite[Section 2.2.3]{wolff_2006}. But it is not clear at present how the SP System would work in processing information of that kind.

\end{itemize}

In general, the SP System works best with SP-symbols representing structures or entities that are significant in human perception. Thus:

\begin{quote}

    ``For the SP System to be effective in the processing of speech or visual images, it seems likely that some kind of preliminary processing will be required to identify low-level perceptual features, such as, in the case of speech, phonemes, formant  ratios  or  formant  transitions,  or,  in  the  case  of  visual  images,  edges,  angles,  colours, luminances or textures.'' \cite[Section 3.3]{sp_extended_overview}.

\end{quote}

In broad terms, there are two ways in which the developing SP Machine may bridge the gap which is likely to exist between information that is drawn relatively directly from the environment and the kinds of structures just mentioned:

\begin{itemize}

    \item As a stop-gap measure, preliminary processing of sensory data may be done by conventional methods to yield the kinds of low-level perceptual entities with which the SP System can operate.

    \item Given the power and generality of the principles that lie at the heart of the SP Theory, it seems likely that, pending some further research, the SP System, without stop-gap measures, will be able to process all kinds of information in a meaningful way, including `raw' sensory data and data that has been encoded using numbers.

\end{itemize}

\section{The Processing of Natural Language}\label{processing_natural_language_section}

A major strength of the SP-multiple-alignment framework is that it lends itself well to the representation and processing of natural language, as described quite fully in \cite[Chapter 5]{wolff_2006} and more briefly in \cite[Section 8]{sp_extended_overview}. Here are some examples:

\begin{itemize}

    \item As can be seen in Figure \ref{parsing_figure} in \ref{patterns_symbols_and_mas_appendix}, and in many other examples in \cite{wolff_2006,sp_extended_overview}, the building of SP-multiple-alignments can achieve the effect of parsing a sentence into its parts and sub-parts.

    \item The SP-multiple-alignment concept exhibits versatility in the representation and processing of several kinds of non-syntactic knowledge (\ref{sp_distinctive_features_advantages_appendix}). Any of these may represent the semantics or meanings of a sentence.

    \item As noted in \ref{sp_distinctive_features_advantages_appendix}, the fact that one relatively simple framework may serve in the representation and processing of diverse kinds of knowledge means that there is potential for their seamless integration in any combination. More specifically in the present context, there is clear potential for the seamless integration of the syntax and semantics of natural language.

    \item Preliminary examples described in \cite[Section 5.7]{wolff_2006} show how this may work in both the semantic interpretation of a simple sentence (\cite[Figure 5.18]{wolff_2006}) and in the production of a sentence from its meanings (\cite[Figure 5.19]{wolff_2006}).

    \item The potential of the SP System for the seamless integration of diverse kinds of knowledge and diverse aspects of intelligence (\ref{sp_distinctive_features_advantages_appendix}) means that there is clear potential for the integration of natural language processing with other aspects of intelligence such as pattern recognition, reasoning, planning, and problem solving.

\end{itemize}

In order to demonstrate parsing via SP-multiple-alignment, the semantic interpretation of a sentence, or the production of a sentence from meanings, it has been necessary to construct appropriate grammars by assembling collections of SP-patterns manually. This has been acceptable in the process of developing the SP-multiple-alignment framework and demonstrating what it can do, but it is likely to be far too slow and prone to error in the further development of natural language processing in the SP System.

To make progress with the processing of natural language in the SP System, it will be necessary to develop this area in conjunction with the development of unsupervised learning in the SP System, as described in the next section.

\section{Unsupervised Learning}\label{unsupervised_learning_development_section}

In the light of observations in the previous section, the discussion of unsupervised learning in this section concentrates mainly on unsupervised learning of the syntax and semantics of natural language. Success with those aspects of natural language is likely to generalise to other kinds of knowledge and, in any case, `semantics' is a broad concept that embraces most kinds of knowledge.

The emphasis on {\em unsupervised} learning in the SP Machine is because of the working hypothesis in this reseaerch that most human learning is achieved without the benefit of any kind of `teacher' or anything equivalent ({\em cf}.~\cite{gold_1967}), and because of a further working hypothesis that other kinds of learning, such as supervised learning, reinforcement learning, learning by being told, and learning by imitation, may be seen as special cases of unsupervised learning \cite[Sections V-A.1 and V-A.2]{sp_autonomous_robots}.

The best place to start in developing unsupervised learning in the SP Computer Model is with two weaknesses in the SP71 model, outlined here:

\begin{quote}

    ``A  limitation  of  the  SP  computer  model  as  it  is  now  is  that  it  cannot learn  intermediate  levels  of  abstraction  in  grammars  (e.g.,  phrases  and  clauses),  and  it  cannot learn the kinds of discontinuous dependencies in natural language syntax that are described in [\cite[Sections 8.1 to 8.3]{sp_extended_overview} and \cite[Section 5.4]{wolff_2006}]. I believe these problems are soluble and that solving them will greatly enhance the capabilities of the system for the unsupervised learning of structure in data ....'' \cite[Section 3.3]{sp_extended_overview}.

\end{quote}

Pending further analysis, the problems with SP71 in discovering structures such as phrases and clauses in the syntax of natural language, and in discovering discontinuous dependencies in that syntax, may both be solved in a manner like what follows.

Since the building of SP-multiple-alignments is an integral part of unsupervised learning in the SP System (\ref{unsupervised_learning_appendix}), and since every stage in the process of building an SP-multiple-alignment yields an encoding of one or more New SP-patterns in terms of the Old SP-patterns in the SP-multiple-alignment, {\em better results may be obtained by applying the learning processes to those encodings as well as to the original SP-patterns}.

Probably, the best way to build up an understanding of what needs to be done is to work with unsegmented samples of relatively simple English-like languages generated by artificial grammars. If or when good insights and good results can be obtained with that kind of material, the system may be tested and refined with more challenging material, as described in the next three subsections.

\subsection{Learning the Syntax of Natural Language}\label{learning_syntax_of_natural_lanuage_section}

Following the steps outlined above, it seems possible now to bring the SP System to a stage of development where it can derive plausible grammars for the syntax of natural language via the application of unsupervised learning to unsegmented textual samples of natural language. This would be a major achievement since it appears that no other system can perform at that level.

Incidentally, for reasons given in \cite[Section 6.2]{sp_extended_overview}, there are likely to be subtle but important differences between grammars derived from natural language texts without meanings and the kinds of grammars that people learn as children, where syntax and semantics are developed in tandem.

It seems likely that success with this task would smooth the path for success with the tasks described next.

\subsection{The Unsupervised Learning of Non-Syntactic Kinds of Knowledge}\label{learning_non-syntactic_knowledge_section}

A key idea in the SP programme of research is that one set of principles applies across the representation and processing of diverse kinds of knowledge.

This idea, which was originally a working hypothesis motivated by Ockham's Razor, is now backed by substance in the form of the concept of SP-multiple-alignment and its versatility in support of diverse aspects of intelligence and its versatility in the representation of diverse forms of knowledge (\ref{sp_distinctive_features_advantages_appendix}). In general, the SP System has potential to be {\em a universal framework for the representation and processing of diverse kinds of knowledge} (UFK) \cite[Section III]{sp_big_data}.

Hence, it is anticipated that if good solutions can be found for the unsupervised learning of syntax, it is likely that, with or without some further development, they would generalise to the learning of non-syntactic, semantic structures.

With the learning of visual structures, it would be necessary for there to be a robust solution to the previously-mentioned problem of finding good full and partial matches between 2D SP-patterns (Section \ref{introduction_of_2d_patterns_section}).

\subsection{Learning Syntactic/Semantic Structures}\label{syntactic_semantics_structures_section}

A prominent feature of the environment in which most young children learn their first language or languages is that, very often, they can hear the speech of adults or older children at the same time as they can see or hear what people are talking about: ``Lunch is on the table'', ``Here's our bus coming'', and so on.

It appears that this kind of environment is needed to enable young children to work out the meanings of words and grammatical forms, and something similar would be needed for experiments with the SP System. But there appears to be a problem:

\begin{quote}

    ``The logician W.~V.~O.~Quine asks us to imagine a linguist studying a newly discovered tribe. A rabbit scurries by, and a native shouts, `Gavagai!' What does gavagai mean? Logically speaking, it needn't be `rabbit.' It could refer to that particular rabbit (Flopsy, for example). It could mean any furry thing, any mammal, or any member of that species of rabbit (say, Oryctolagus cuniculus), or any member of that variety of that species (say, chinchilla rabbit). It could mean scurrying rabbit, scurrying thing, rabbit plus the ground it scurries upon, or scurrying in general....~Somehow a baby must intuit the correct meaning of a word and avoid the mind-boggling number of logically impeccable alternatives. It is an example of a more general problem that Quine calls `the scandal of induction,' which applies to scientists and children alike: how can they be so successful at observing a finite set of events and making some correct generalization about all future events of that sort, rejecting an infinite number of false generalizations that are also consistent with the original observations?'' \cite[pp.~147--148]{pinker_1995}.

\end{quote}

Without wishing to trivialise this problem, it appears that the SP System has potential to provide a good solution:

\begin{itemize}

    \item By searching for recurrent patterns and attaching more weight to patterns that occur frequently and less weight to patterns that are rare, the SP System naturally homes in on the stronger correlations---such as the correlation between the word ``lunch'' and what people eat at lunchtime---and discards weaker correlations---such as the correlation between the word ``lunch'' and whether or not it is raining outside.

    \item The problem of how to generalise from what one has learned, without either under-generalisation or over-generalisation, is discussed in \cite[Section 5.3]{sp_extended_overview}\footnote{Under-generalisation was overlooked in this account but the avoidance of under-generalisations is achieved in exactly the same way as the avoidance of over-generalisations.} and also in \cite[Section V-H]{sp_alternatives}.

        In brief, learning with the SP System from a body of data, {\bf I}, means compressing {\bf I} as much as possible and then splitting the resulting body of data into two parts: a {\em grammar} which contains a single copy of each recurrent SP-pattern in {\bf I}; and an {\em encoding} of {\bf I} which contains non-recurrent parts of {\bf I} and brief codes or references to recurrent SP-patterns in {\bf I}.\footnote{Notice that recurrent SP-patterns in {\bf I} may be abstract patterns as well as concrete patterns.}

        Evidence to date shows that with learning of this kind, the grammar generalises in a way that native speakers judge to be `correct' and avoids what intuitively such people would judge to be under-generalisation or over-generalisation.\footnote{See also the discussion of generalisation in \cite{wolff_1988}.}

\end{itemize}

Assuming that progress can be made along the lines outlined here, another avenue that may be explored later is how children may learn the meanings of things by making inferences from their existing knowledge. For example, if a child that does not know the word ``plantain'' hears someone say something like: ``I had plantain for lunch today. It was delicious---more tasty than potato but not overpowering'', he or she may infer quite reasonably that plantain is something that one can eat and that it is probably something starchy, like potato.''

In this area of research---how the syntax and semantics of natural language may be learned together---it will be necessary to provide input to the SP System that is comparable with that described above.

Also, it will probably be necessary for the SP System to have been generalised to work with two-dimensional SP-patterns (Section \ref{introduction_of_2d_patterns_section}), so that syntactic knowledge may be shown alongside semantic knowledge, with associations between them (see also discussions in \cite[Sections V-G, V-H, and V-I, and Appendix C]{sp_autonomous_robots}).

\subsection{Learning and Use of Causal Relations}

Since concepts of causation and how they may be learned have been the subject of much research and discussion by, for example, Judea Pearl \cite{pearl_2000,pearl_mackenzie_2018}, it will not be possible here to do more than make a few tentative remarks and suggestions.

In view of the versatility of the SP System in aspects of intelligence and in the representation of knowledge (\ref{sp_distinctive_features_advantages_appendix}), and in view of the central role of information compression in the workings of the SP System (\ref{icmup_appendix} and \ref{multiple_alignments_appendix}), it seems likely that information compression will also have an important explanatory role in the representation and learning of causal relations.

Although superficially there may seem to be little in common between information compression and causal relations, the connection may make more sense in the light of the intimate connection that is known to exist between information compression and concepts of correlation, inference, and probability (\ref{correlation_inference_probability_appendix}).

But it is well known that finding a correlation between two things, such as smoking and cancer, is not the same as demonstrating a causal relation. A useful example in this connection is that two radios, R1 and R2, that are tuned to the same transmitter, T, will produce sounds that are perfectly correlated, but it is clear that neither radio has a causal connection with the other. Of course, it is T which is the real driver of both R1 and R2.

Examples like these seem, superficially, to invalidate any scheme that attempts to explain the learning and representation of causal relations in terms of correlation and closely-related concepts of information compression.

But: a full understanding of the relationships amongst T, R1, and R2 may be regarded as a `deeper' understanding than one that merely sees the correlation between R1 and R2; and that intuitive concept of `depth' of understanding may turn out to be expressible in terms of levels of information compression; then, it may prove feasible in the SP System to learn causal relations via information compression, in much the same way that other kinds of knowledge may be learned via compression of information.

\subsection{Learning Language From Speech}

Although, to a large extent, children learn their first language or languages by hearing people talk, learning a language via speech has in this paper been left to last because it is more challenging than, for example, learning grammars from text.

The main reason for the difficulty is uncertainties about the discovery of low-level features in speech, as discussed in Section \ref{low_level_features_in_speech_and_vision_section}. Solutions in that area will facilitate the learning of language from speech.

\subsection{Learning from Data in Two and Three Dimensions}

As noted in Section \ref{introduction_of_2d_patterns_section} a significant challenge will be learning from SP-patterns in two dimensions, and as noted in Section \ref{four_or_more_dimensions_section}, an important application of the SP System will be learning, from videos or the like, 2D and 3D structures that change across the time dimension, and with sounds including speech associated with vision.

This last kind of learning approximates what we all do in our learning from birth onwards. It may be seen as a `holy grail' for learning---something which will be necessary but not sufficient to realise the full potential of the SP System.

\section{Computer Vision}\label{computer_vision_section}

The generalisation of the SP Machine to work with 2D SP-patterns (Section \ref{introduction_of_2d_patterns_section}) is a likely prerequisite for the development of the SP System for computer vision and as a model of human vision, as described in \cite{sp_vision}.

An important part of this development will be the unsupervised learning or discovery of objects and classes, as discussed in \cite[Section 5]{sp_vision} and Section \ref{learning_non-syntactic_knowledge_section}, below. This will include work on the discovery of low-level features in images, as discussed in Section \ref{low_level_features_in_speech_and_vision_section}.

\section{Processing Numbers}\label{processing_numbers_section}

In view of evidence that mathematics may be seen as a set of techniques for compression of information, and their application \cite{sp_maths_mystery}, and bearing in mind that the SP System works entirely via the compression of information, and in view of the potential of the SP System as a vehicle for `computing' (\cite[Chapter 4]{wolff_2006}, \cite[Section 6.6]{sp_benefits_apps}, \cite{sp_software_engineering}), and as a vehicle for mathematics \cite[Chapter 10]{wolff_2006}, there is clear potential in the SP Machine for the performance of arithmetic and related computations. But the SP Computer Model does not yet handle numbers in a satisfactory way.

Does this matter? In terms of short-term practicalities, the answer is ``no''---because it would be easy enough to provide some kind of maths co-processor within the SP Machine that would handle numbers and arithmetic operations very well. But in terms of the long-term objectives of the SP programme of research---creating a system with a favourable combination of conceptual {\em simplicity} with descriptive and explanatory {\em power}---the use of a maths co-processor would be an {\em ad hoc} development, much like a software `hack' without good design.

We believe that, as far as possible, the long-term objectives of the SP research should be maintained (Section \ref{strong_focus_section}). This strategy has yielded substantial dividends already and we believe there is much more to come.

But pending more mature capabilities for arithmetic in the SP Machine, there may be a case for using an existing maths coprocessor or similar facility as an adjunct to the SP Machine, much as is done in many database management systems today.

\section{SP-Neural}\label{sp-neural_section}

At present, the SP-Neural part of the SP Theory (\ref{sp-neural_appendix}) exists only as a conceptual model, described in some detail in \cite{spneural_2016}.

To flesh out this conceptual model, it is really necessary to create a computational version of it. As with the SP71 computer model, this will help to guard against vagueness in the theory, it will serve as a means of testing ideas to see whether or not they work as anticipated, and it will be a means of demonstrating what the model can do.

\begin{figure*}[!htbp]
\centering
\includegraphics[width=0.7\textwidth]{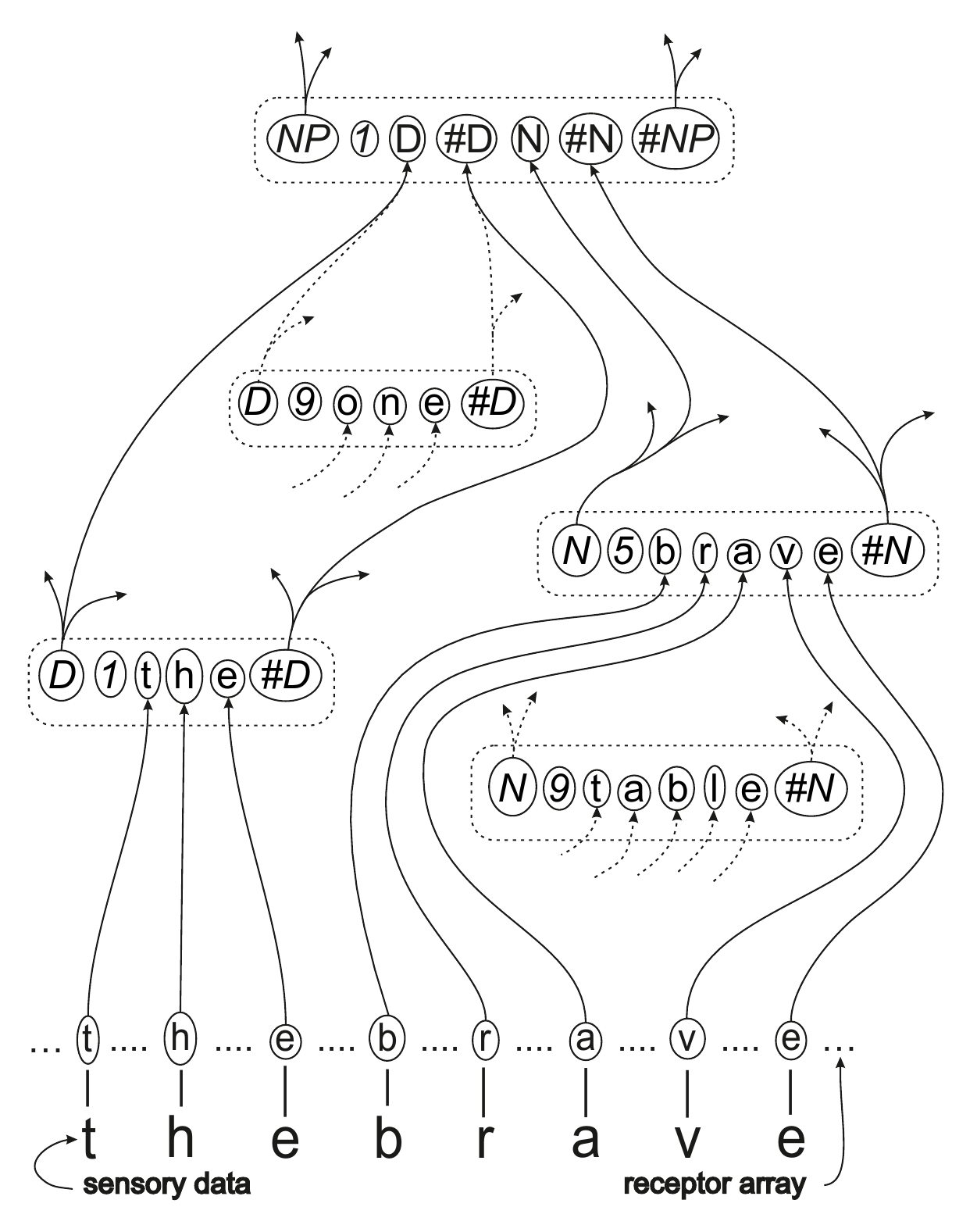}
\caption{A schematic representation of part of the process of building an SP-multiple-alignment in SP-Neural, as discussed in \cite[Section 4]{spneural_2016}. Each broken-line rectangle with rounded corners represents a {\em pattern assembly}---corresponding to a `pattern' in the main SP Theory; each character or group of characters enclosed in a solid-line ellipse represents a {\em neural symbol} corresponding to a `SP-symbol' in the main SP Theory; the lines between pattern assemblies represent nerve fibres with arrows showing the direction in which impulses travel; neural symbols are mainly symbols from linguistics such as `\texttt{NP}' meaning `noun phrase, `\texttt{D}' meaning a `determiner', `\texttt{\#D}' meaning the end of a determiner, `\texttt{\#NP}' meaning the end of a noun phrase, and so on. Reproduced with permission from Figure 3 in \cite{spneural_2016}.}
\label{the_brave_neural_figure}
\end{figure*}

The best place to start would be to create a computer model of the building of an SP-multiple-alignment in SP-Neural, as discussed in \cite[Section 4]{spneural_2016} and illustrated schematically in Figure \ref{the_brave_neural_figure}.

The development of unsupervised learning in the computer model of SP-Neural is probably best left until later, if or when there is a satisfactory SP-Neural model of the building of the neural equivalent of SP-multiple-alignments. This is because the building of SP-multiple-alignments is an important part of unsupervised learning in the SP System. There are some relevant remarks about unsupervised learning in Section \ref{unsupervised_learning_in_sp_neural_section}.

In Figure \ref{the_brave_neural_figure}, each alphanumeric symbol surrounded by a solid-line envelope represents a {\em neural symbol} (\ref{sp-neural_appendix}), and each group of neural symbols surrounded by a broken-line envelope represents a {\em pattern assembly} (\ref{sp-neural_appendix}, \cite[Section 4.2]{spneural_2016}).

In the figure, nerve fibre connections between neural symbols are represented with lines which may be solid or broken. The solid lines represent neural connections in part of the SP-multiple-alignment shown in \cite[Figure 2]{spneural_2016}, while the broken lines represent other neural connections that are not part of that SP-multiple-alignment.

\subsection{Inhibition}\label{inhibition_section}

This section considers the potential role of inhibitory processes in the new computer model, starting with some background.

\subsubsection{Background}

In \cite[Section 9]{spneural_2016}, there is a brief review of what appears to be a substantial role for inhibitory signals between neurons in the workings of brains and nervous systems. A tentative general rule from this research is that {\em When, in neural processing, two or more signals} [that interact] {\em are the same, they tend to inhibit each other, and when they are different, they don't} \cite[p.~19]{spneural_2016}, with the implication that ``The overall effect should be to detect redundancy in information and to reduce it, whilst retaining non-redundant information, in accordance with the central principle in the SP Theory---that much of computing and cognition may, to a large extent, be understood as information compression.'' ({\em ibid}.).

\begin{figure}[!htbp]
\centering
\includegraphics[width=0.4\textwidth]{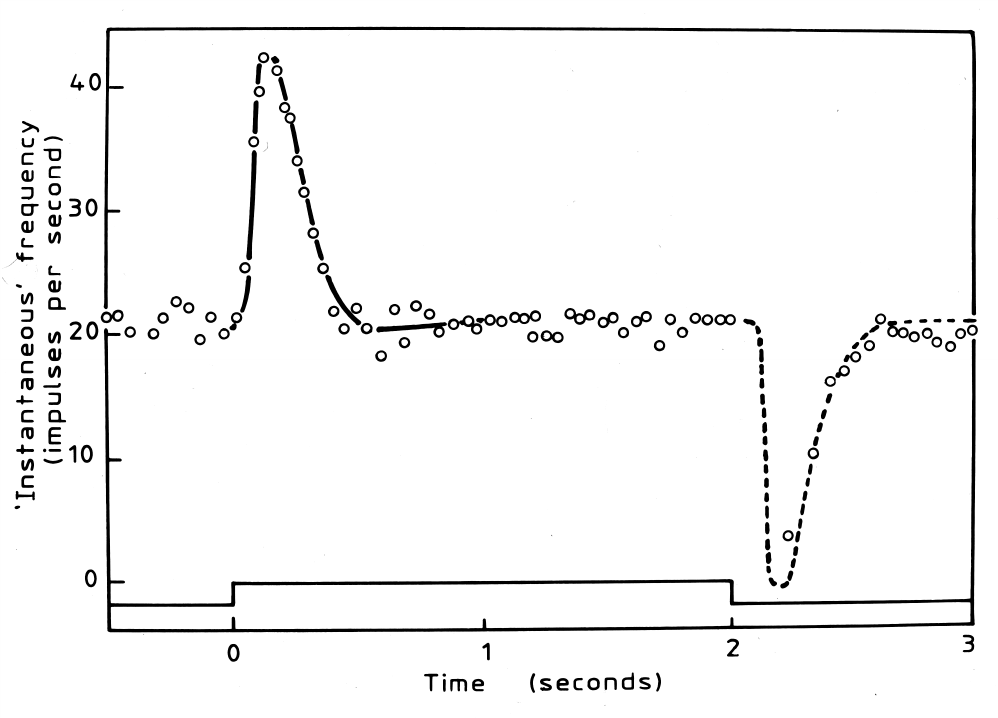}
\caption{Variation in the rate of firing of a single ommatidium of the eye of a horseshoe crab in response to changing levels of illumination. Reproduced from \protect\cite[Figure 16]{ratliff_etal_1963}, with permission from the Optical Society of America.}
\label{limulus_figure}
\end{figure}

A related idea is that the rate of firing of neurons correlates only indirectly with the strength of stimulation, and that when stimulation is steady, neurons typically fire at a constant rate which is the same regardless of the strength of the stimulation. This can be seen in Figure \ref{limulus_figure} which shows variations in the rate of firing of a sensory cell ({\em ommatidium}) in the eye of the horseshoe crab ({\em Limulus}) in response to variations in the intensity of the light that is falling on it.

On the left of the figure, where the light level is low, the ommatidium fires at a steady rate of about 20 impulses per second. When a light is switched on, there is sharp upswing in the rate of firing, but this soon settles back to the same 20 impulses per second rate as before. When the light is switched off, there is a sharp downswing in the rate of firing followed by a rapid return to the previous steady rate.

This pattern of responses can be explained via inhibitory connections which dampen down responses when stimulation is steady \cite{ratliff_etal_1963,von_bekesy_1967}.\footnote{Although the two sources that have been referenced have authority in this area, they do not explain the principles very well. A clearer account may be found in \cite[pp.~65--75]{lindsay_norman_1972}.} Similar effects occur in the spatial dimension and it appears that similar explanations apply.

\subsubsection{Developments}

In the building of an SP-multiple-alignment, modelling such things as the parsing of a sentence or the recognition of a pattern, things are a bit different from what has been described in the previous subsection. Instead of trying to detect and remove redundancy between neighbouring elements in a sequential or spatial pattern, the aim is to detect redundancies between two independent patterns, such as a New SP-pattern (as, for example, `\texttt{I N F O R M A T I O N}') and an Old SP-pattern (such as `\texttt{N 6 I N F O R M A T I O N \#N}'), or between two partial SP-multiple-alignments at later stages of processing.

\begin{figure}[!htbp]
\centering
\includegraphics[width=0.4\textwidth]{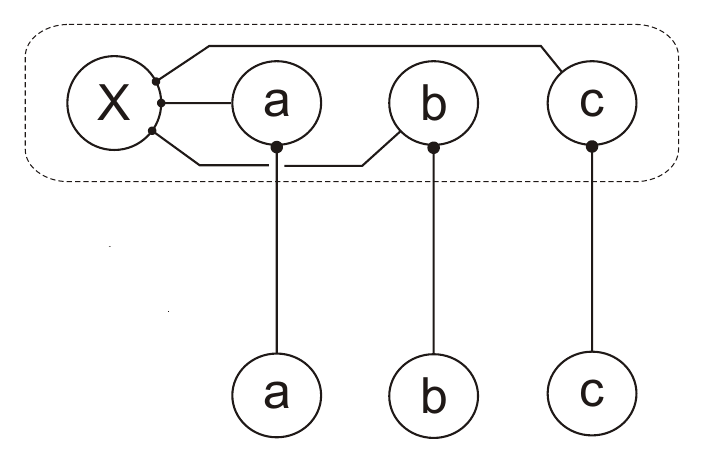}
\caption{A schematic representation of one possible neural scheme for detecting redundancies between an incoming pattern (`\texttt{a b c}'), shown at the bottom of the figure and a stored pattern that matches it, shown above. {\em Key:} a letter inside an unbroken envelope represents a neural symbol; a group of neural symbols surrounded by a broken-line envelope represents a pattern assembly; lines between neural symbols ending in a blob (a filled circle) represent nerve fibres carrying signals towards the blob, with an inhibitory impact on the neural symbol that is receiving the signals.}
\label{inhibition_figure}
\end{figure}

Bearing in mind the putative idea that the role of inhibition is to detect and remove redundancies in information, a possible scheme is shown in Figure \ref{inhibition_figure}, with a key to the meanings of SP-symbols in the caption to the figure.

This is how the scheme may work:

\begin{itemize}

    \item In accordance with what was said about the results of neural processing shown in Figure \ref{limulus_figure}, it is envisaged that, without incoming signals, the four neural symbols shown at the top of the figure will fire at a steady intermediate rate.

    \item If a signal comes in from the neural symbol marked `\texttt{a}' at the bottom of the figure, this will have an inhibitory effect on the neural symbol marked `\texttt{a}' at the top of the figure. Likewise for the other two neural pairings, `\texttt{b}' with `\texttt{b}', and `\texttt{c}' with `\texttt{c}'.

    \item If all three of `\texttt{a}', `\texttt{b}', and `\texttt{c}', in the pattern assembly have been inhibited in this way, there will be reduced inhibitory signals transmitted to the ``identifier'' (ID) neural symbol marked `\texttt{X}' in the figure.

    \item Since the ID neural symbol is receiving reduced inhibitory signals, its rate of firing will increase. This may be interpreted as recognition of the pattern `\texttt{a b c}'. That increase in the rate of firing may be transmitted to higher-level pattern assemblies that contain `\texttt{X}' as a ``reference'' to the pattern `\texttt{a b c}'. This may lead to similar processes of recognition at higher levels.

\end{itemize}

A possible weakness in the account just given is that, in line with what was said about Figure \ref{limulus_figure}, it may be supposed that the rate of firing of the `\texttt{X}' neural symbol would quickly settle back to the average rate of firing, but there is nothing to ensure that that would happen. Issues like this may be clarified with a working model.

What about the ordering of neural symbols? With the scheme as just described, the pattern assembly `\texttt{X a b c}' would respond as strongly to an input pattern `\texttt{b c a}' as it would to the input pattern `\texttt{a b c}'. But it looks as if the concept of a {\em typographic pattern}, provides an answer. As was quoted in \cite[Section 5.4]{spneural_2016}:

\begin{quote}

    ``Receptors in the retina and body surface are organized as two-dimensional sheets, and those of the cochlea form a one-dimensional line along the basilar membrane. Receptors in these organs communicate with ganglion cells and those ganglion cells with central neurons in a strictly ordered fashion, such that relationships with neighbors are maintained throughout. {\bf This type of pattern, in which neurons positioned side by side in one region communicate with neurons positioned side-by-side in the next region, is called a {\em typographic pattern}}.'' \cite[p.~504]{squire_etal_2013}, italic emphasis in the original, bold face added.

\end{quote}

What this concept of a `typographic pattern' would mean for the workings of the scheme shown in Figure \ref{inhibition_figure} is that the neural symbol `\texttt{X}' is a detector for the pattern `\texttt{a b c}' and not for `\texttt{b c a}', `\texttt{c b a}', and so on, because it only ever receives signals in the order `\texttt{a b c}'.

\subsection{Implementation of the SP-Neural Computer Model}\label{implementation_of_sp_neural_section}

The SP-Neural computer model may be developed with these main features:

\begin{itemize}

    \item {\em The C++ computer language}. To be compatible with the existing SP Computer Model (SP71), the model is probably best implemented using the C++ computer language.

    \item {\em Neural symbols as a class}. Neural symbols would be implemented as a C++ class. For each instance of the class, this would provide the following features:

    \begin{itemize}

        \item {\em Inhibitory connections between neurons}. The paths for inhibitory messages to one or more neural symbols would be nerve-fibre connections like the one between neural symbol `\texttt{a}' near the bottom of Figure \ref{inhibition_figure} and neural symbol `\texttt{a}' near the top, and likewise for the two other pairs of neural symbols in the figure.

        \item {\em Indexing of symbols that match each other}. If the nerve-fibre connections just mentioned are only ever between a pair of symbols that match each other, such as, for example, between `\texttt{N}' as the ID neural symbol in a pattern assembly like `\texttt{N d o g}' (the word `dog' belongs in the class `noun'), and `\texttt{N}' as part of the contents of a pattern assembly like `\texttt{S N V}' (a sentence in its simplest form is a noun followed by a verb), then the list of connections in each neural symbol is, in effect, an index of matching symbols, much like the index proposed for the abstract part of the SP Theory, described in Section \ref{indexing_for_speed_section}.

            This line of thinking suggests that many neural connections in the brain, perhaps all of them, are, in effect, direct connections between neural symbols with a role that is very much like that of an index in any ordinary database management system, or the very large indices that are maintained in search engines for the internet.

        \item {\em Inhibition of ID neural symbols}. Within each pattern assembly, each neural symbol would have the means of sending an inhibitory message to ID neural symbol at the far left of the pattern assembly, as shown in Figure \ref{inhibition_figure}. Notice that any one neuron can only ever be in one pattern assembly, or none. Hence, any one neuron only ever needs to be able to send inhibitory messages to an ID neural symbol in the pattern assembly of which it is a part.

        \item {\em Subclasses of the neural symbol class}. There may be a case for providing two subclasses of the neural symbol class: one for {\em C neural symbols} (each of which may be part of the contents of a pattern assembly) and one for {\em ID neural symbols} (each of which may serve as an identifier for a pattern assembly). These features of SP-Neural are described in \cite[Section 11.3.2]{wolff_2006}.\footnote{The term {\em C-neuron} (meaning ``C neural symbol''), introduced in \cite[Section 11.3.2]{wolff_2006}, and the term {\em ID-neuron} (meaning ``ID neural symbol''), introduced in the same place, should probably be dropped. This is because neural symbols are not necessarily single neurons---they may be small clusters of neurons.}

    \end{itemize}

    \item {\em Pattern assemblies as a class}. Pattern assemblies would also be implemented as a C++ class. For each instance of the class, it would provide the means of storing zero or more ID-neural-symbols and zero or more C-neural-symbols, perhaps using linked lists. It may also store other information such as the number of times the pattern assembly has been recognised in a given set of New SP-patterns.

\end{itemize}

\subsection{Unsupervised Learning in SP-Neural}\label{unsupervised_learning_in_sp_neural_section}

How unsupervised learning may be achieved in SP-Neural is considered mainly in \cite[Section 10]{spneural_2016} with some related observations in Section 11 of that paper. Here, there are no new insights to be added to that account except to underscore the point that:

\begin{quote}

    ``...~in any or all of short-term memory, working memory, and long-term memory, SP-Neural may achieve the necessary speed in the creation of new structures, combined with versatility in the representation and processing of diverse kinds of knowledge, by the switching on and off of synapses in pre-established neural structures and their inter-connections ...'' and that this would be ``somewhat like the way in which an `uncommitted logic array' (ULA) may, via small modifications, be made to function like any one of a wide variety of `application-specific integrated circuits' (ASICs), or how a `field-programmable gate array' (FPGA) may be programmed to function like any one of a wide variety of integrated circuits.'' \cite[Section 11]{spneural_2016}.

\end{quote}

\section{New Hardware}\label{new_hardware_section}

As was mentioned in the Introduction, there may, at some stage, be a case for the development of new hardware, dedicated to structures and processes in the SP System, and with potential for gains in efficiency and performance (\cite[Section IX]{sp_big_data}, \cite[Section III]{sp_autonomous_robots}).

Since searching for matching SP-patterns is an important part of how the SP System works, gains in efficiency may be achieved by concentrating search where good results are most likely to be found: ``If we want to find some strawberry jam, our search is more likely to be successful in a supermarket than it would be in a hardware shop or a car-sales showroom.'' \cite[Section IX-A.2]{sp_big_data}, and the statistical knowledge in the system flows directly from the central role of information compression in the workings of the SP System, and from the intimate relationship that is known to exist between information compression and concepts of correlation, inference and probability (\ref{correlation_inference_probability_appendix}).

Gains in efficiency from the use of indices (Section \ref{indexing_for_speed_section}) may be at least partly via the encoding of statistical aspects of the data that are being processed.

New hardware may be developed for either or both of the two versions of the SP System: the abstract version embodied in the SP71 computer model, and SP-Neural as outlined in \ref{sp-neural_appendix}, and discussed in Section \ref{sp-neural_section}.

\section{Development of Applications of the SP Machine}\label{devt_apps_sp_machine_section}

Sources of information about potential applications of the SP Machine are detailed in \ref{potential_benefits_applications_appendix}. Each one of these potential applications may be the subject of one or more research projects in its own right.

\section{Applications That May Be Realised Quite Soon}\label{apps_on_short_timescales_section}

In case practical applications of the SP Machine seem too distant to be supported, some of them may be realised quite soon. Here, in brief, are some examples:

\begin{itemize}

    \item {\em The SP Machine as an intelligent database system}. With the application of parallel processing and indexing (Section \ref{high_parallel_sp_machine_section}) and the creation of a ``friendly'' user interface (Section \ref{user_friendly_user_interface_section}), the SP Machine should begin to be useful as an intelligent database system, with several advantages over ordinary database systems, as described in \cite{wolff_sp_intelligent_database}.

    \item {\em Information compression}. Since information compression is central in how the SP System works, the SP Machine is likely to prove useful in situations where compression of information is the main requirement. It may, for example, be a useful means of reducing the size of ``big data'', with corresponding benefits in storage and transmission of those data \cite[Section VII]{sp_big_data}.

    \item {\em Several kinds of reasoning}. Most of the several kinds of reasoning described in \cite[Chapter 7]{wolff_2006} will be available immediately in the highly-parallel version of the SP Machine. Because they all flow from one simple framework---SP-multiple-alignment---they may be applied in any combination (meaning all the several kinds of reasoning or any subset of them), and with seamless integration amongst them.

    \item {\em Model-based coding for the efficient transmission of information}. The concept of model-based coding was described by John Pierce in 1961 like this:

        \begin{quote}

            ``Imagine that we had at the receiver a sort of rubbery model of a human face. Or we might have a description of such a model stored in the memory of a huge electronic computer. First, the transmitter would have to look at the face to be transmitted and `make up' the model at the receiver in shape and tint. The transmitter would also have to note the sources of light and reproduce these in intensity and direction at the receiver. Then, as the person before the transmitter talked, the transmitter would have to follow the movements of his eyes, lips and jaws, and other muscular movements and transmit these so that the model at the receiver could do likewise.'' \cite[pp.~139--140]{pierce_1961}.

        \end{quote}

        At the time this was written, it would have been impossibly difficult to make things work as described. But it seems likely that, with the SP System, it will become feasible fairly soon in a simple but effective form, and it will improve progressively with improvements in unsupervised learning in the SP System.

\end{itemize}

\section{Conclusion}

This paper describes a roadmap for the development of a mature version of the SP Machine, starting with the SP71 computer model. After an overview of the strategy adopted for this programme of research, there is a description of the main tasks that are to be tackled, probably in the following order:

\begin{enumerate}

    \item Starting with the SP71 computer model, create a highly-parallel version of the SP Machine as a software virtual machine hosted on a high-performance computer. An interesting alternative would be to create the SP Machine as a software virtual machine driven by highly-parallel search processes in any of the leading search engines.

    \item Create a user interface for the SP Machine that is easy to use.

    \item Closely related to the previous task is the development of the SP System's existing strengths in the visualisation of the structure of stored knowledge and in the provision of an audit trail for all processing.

    \item As the basis for such things as computer vision, the SP Machine will need to be generalised to work with SP-patterns in two dimensions as well as the 1D SP-patterns of the SP71 model.

    \item The generalisation just mentioned would include the development of processes for the discovery of good full and partial matches between 2D SP-patterns (and between a 2D SP-pattern and a 1D SP-pattern) and for the display of SP-multiple-alignments comprising two or more 2D SP-patterns (with or without 1D SP-patterns).

    \item An investigation is needed into whether or how the SP Machine, with or without some modification, may discover low-level features in speech and visual images, of the kind that appear to be significant in human perception.

    \item Existing strengths of the SP System in the processing of natural language may be developed towards the goal of creating an SP Machine that can understand natural language and produce natural language from meanings.

        This is likely to be a major project or programme of research, with several stages, in conjunction with the development of unsupervised learning (next bullet point).

        Likely stages include: experiments with the unsupervised learning of artificial languages: unsupervised learning of plausible grammars from unsegmented samples of natural language text; unsupervised learning of plausible structures for various kinds of non-syntactic knowledge; unsupervised learning of syntactic/semantic structures; and unsupervised learning from speech.

    \item Residual problems with the existing processes for unsupervised learning need to be solved. These weaknesses are a failure of the system to discover grammatical structures with intermediate levels of abstraction, and failure to discover discontinuous dependencies in such structures.

        A solution to these problems will clear the path for the development of unsupervised learning in conjunction with the processing of natural language, as described above.

    \item Existing strengths of the SP System in pattern recognition may be developed for computer vision, guided by the insights in \cite{sp_vision}.

    \item On the strength of evidence to date, it seems likely that the representation of numbers and the performance of arithmetic processes may be accommodated within the SP framework. Pending success in those areas, stop-gap solutions may be employed where needed.

    \item Somewhat independent of other developments described in this paper, would be the development of a computer model of SP-Neural, drawing on the information and insights described in \cite{spneural_2016} and perhaps also in \cite[Chapter 11]{wolff_2006}.

        An important part of this development would be to see whether or how inhibitory processes have the important role that seems likely from evidence that is available now, and, if not, to provide some reasons why.

    \item At some stage when the SP Machine is relatively mature, it is envisaged that new hardware would be developed, mainly to exploit opportunities to increase the efficiency of computations, most notably by taking advantage of statistical information that the SP System gathers as a by-product of how it works.

\end{enumerate}

There is potential for the SP Machine to be applied on relatively short timescales in such areas of application as information storage and retrieval, information compression, several kinds of reasoning, and model-based coding for the efficient transmission of information.

\section*{Acknowledgements}

We are grateful to anonymous reviewers for many helpful suggestions and corrections.

\appendix

\section{Outline of the SP System}\label{outline_of_sp_system_appendix}

To help ensure that this paper is free standing, the SP System is described in outline here with enough detail to make the rest of the paper intelligible.

The SP System is described most fully in \cite{wolff_2006} and more briefly in \cite{sp_extended_overview}.

\begin{figure}[!htbp]
\centering
\includegraphics[width=0.4\textwidth]{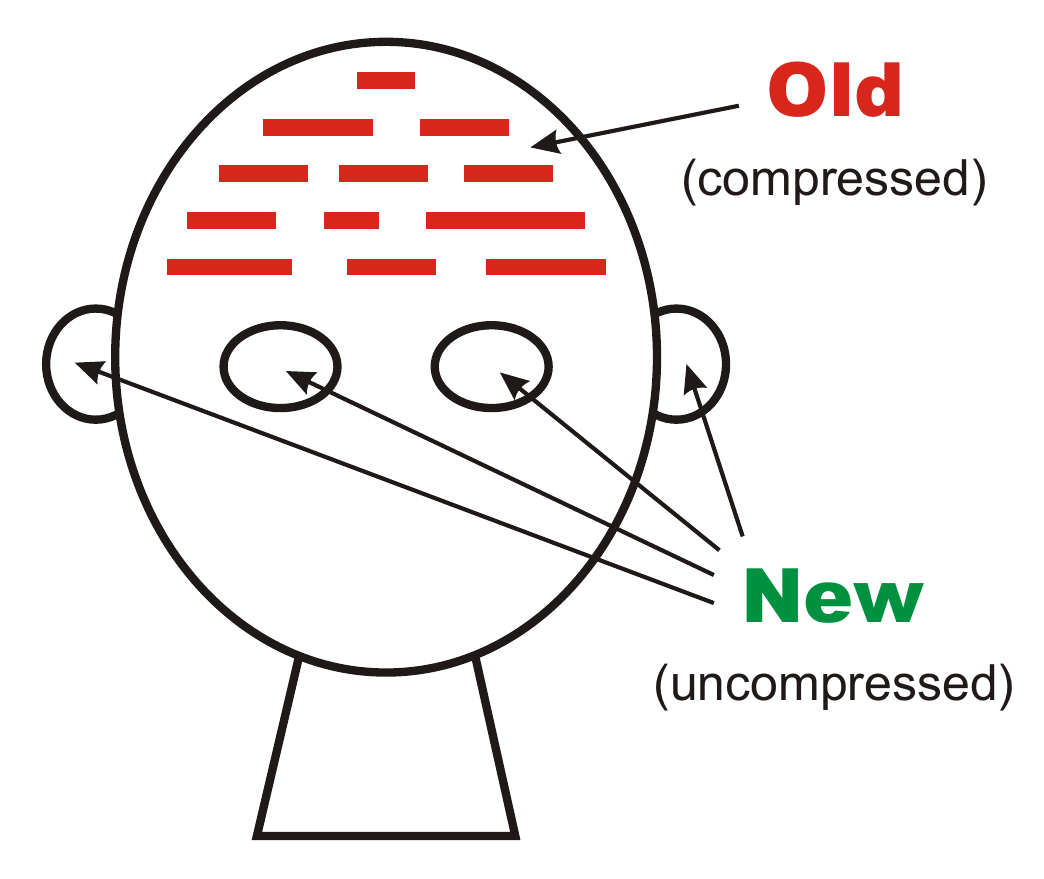}
\caption{Schematic representation of the SP System from an `input' perspective. Reproduced with permission from Figure 1 in \cite{sp_extended_overview}.}
\label{sp_input_perspective_figure}
\end{figure}

The SP Theory is conceived as a brain-like system as shown schematically in Figure \ref{sp_input_perspective_figure}. The system receives {\em New} information via its senses and stores some or all of it in compressed form as {\em Old} information.

\subsection{Information Compression Via the Matching and Unification of Patterns}\label{icmup_appendix}

Information compression is central in the SP System because of substantial evidence for the importance of information compression in the workings of brains and nervous systems described in \cite{sp_compression}.

Since people often ask what the meaning is of the name ``SP''. it is short for {\em Simplicity} and {\em Power}. This is because information compression may be seen to be a process of maximising the {\em simplicity} of a body of information, {\bf I}, by extracting redundancy from {\bf I}, whilst retaining as much as possible of its non-redundant descriptive {\em power}.

As the title of this subsection suggests, information compression in the SP System is not any kind of information compression, it is information compression that is achieved via a search for patterns that match each other and the merging or `unification' of patterns that are the same. This orientation has been adopted because it seems to provide a better handle on possible mechanisms for the compression of information in natural or artificial systems than do the more mathematically-oriented approaches to information compression.

\subsection{Information Compression and Concepts of Correlation, Inference and Probability}\label{correlation_inference_probability_appendix}

It has been known for some time that there is an intimate connection between information compression and concepts of correlation, inference and probability \cite{shannon_weaver_1949,solomonoff_1964,solomonoff_1964,li_vitanyi_2014}. Hence, dedication of the SP System to information compression means that, in the SP System, it is relatively straightforward to make inferences and to calculate associated probabilities.

\subsection{SP-Multiple-Alignments}\label{multiple_alignments_appendix}

In the SP System, information compression is achieved largely via the concept of {\em SP-multiple-alignment}, a concept that has been adapted from the concept of `multiple sequence alignment' in bioinformatics.

As illustrated in Figure \ref{DNA_figure}, a multiple sequence alignment in bioinformatics is an arrangement of two or more DNA sequences or amino-acid sequences so that, by judicious `stretching' of sequences in a computer, symbols that match each other are brought into line. A `good' multiple sequence alignment is one in which a relatively large number of matching symbols have been aligned.

\begin{figure*}[!htbp]
\fontsize{10.00pt}{12.00pt}
\centering
{\bf
\begin{BVerbatim}
  G G A     G     C A G G G A G G A     T G     G   G G A
  | | |     |     | | | | | | | | |     | |     |   | | |
  G G | G   G C C C A G G G A G G A     | G G C G   G G A
  | | |     | | | | | | | | | | | |     | |     |   | | |
A | G A C T G C C C A G G G | G G | G C T G     G A | G A
  | | |           | | | | | | | | |   |   |     |   | | |
  G G A A         | A G G G A G G A   | A G     G   G G A
  | |   |         | | | | | | | |     |   |     |   | | |
  G G C A         C A G G G A G G     C   G     G   G G A
\end{BVerbatim}
}
\caption{A `good' multiple alignment amongst five DNA sequences. Reproduced with permission from Figure 3.1 in \cite{wolff_2006}.}
\label{DNA_figure}
\end{figure*}

Because there is normally an astronomically large number of alternative ways in which multiple sequence alignments may be arranged, heuristic methods are needed to build multiple sequence alignments in stages, discarding all but the best few partial multiple sequence alignments at the end of each stage. With such methods, it is not normally possible to guarantee that a theoretically ideal multiple sequence alignment has been found, but it is normally possible to find multiple sequence alignments that are `reasonably good'.

The following subsections describe how the multiple sequence alignment concept has been adapted to become the concept of SP-multiple-alignment.

\subsection{SP-Patterns, SP-Symbols, and SP-Multiple-Alignments in the SP System}\label{patterns_symbols_and_mas_appendix}

In the SP System, {\em all} kinds of knowledge are represented with arrays of atomic {\em SP-symbols} in one or two dimensions, termed {\em SP-patterns}. Of course, a 1D SP-pattern is the same as a sequence in bioinformatics, but in the SP System the concept of an SP-pattern includes both 1D and 2D arrays of symbols. Although, at present, the SP Computer Model works only with 1D SP-patterns, it is envisaged that it will be developed to work with 2D SP-patterns as well (Section \ref{high_parallel_sp_machine_section}).

There may also be a case for introducing unordered SP-patterns (sets) in which the order of the symbols is not significant---but that effect may be achieved via the unordered nature of sets of New SP-patterns ({\em cf}., \cite[Section 3.4]{wolff_medical_diagnosis}).

In themselves, SP-patterns are not very expressive. But within the SP-multiple-alignment framework they become very versatile as a means of representing diverse kinds of knowledge \cite[Appendix B.1]{sp_software_engineering}.

The main change to the multiple sequence alignment concept in the SP System is that, in each SP-multiple-alignment, one or more of the sequences (often only one) is a New SP-pattern and the rest are Old, and a `good' SP-multiple-alignment is one in which the New information may be encoded economically in terms of the Old information, as described in \cite[Section 2.5]{wolff_2006} and in \cite[Section 4.1]{sp_extended_overview}.

\begin{figure*}[!htbp]
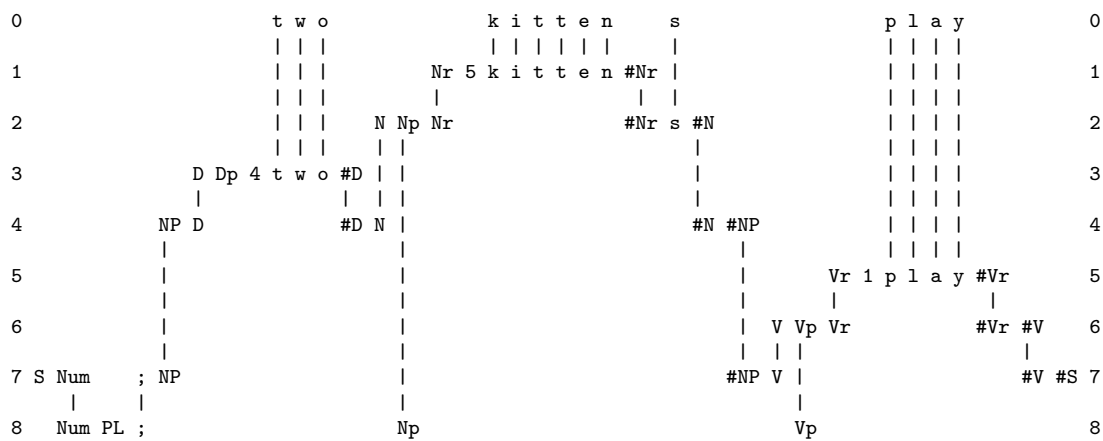

\fontsize{08.00pt}{09.60pt}
\centering
{\bf
\begin{BVerbatim}
0                      t w o              k i t t e n     s                  p l a y           0
                       | | |              | | | | | |     |                  | | | |
1                      | | |         Nr 5 k i t t e n #Nr |                  | | | |           1
                       | | |         |                 |  |                  | | | |
2                      | | |    N Np Nr               #Nr s #N               | | | |           2
                       | | |    | |                         |                | | | |
3               D Dp 4 t w o #D | |                         |                | | | |           3
                |            |  | |                         |                | | | |
4            NP D            #D N |                         #N #NP           | | | |           4
             |                    |                             |            | | | |
5            |                    |                             |       Vr 1 p l a y #Vr       5
             |                    |                             |       |             |
6            |                    |                             |  V Vp Vr           #Vr #V    6
             |                    |                             |  | |                   |
7 S Num    ; NP                   |                            #NP V |                   #V #S 7
     |     |                      |                                  |
8   Num PL ;                      Np                                 Vp                        8
\end{BVerbatim}
}
\caption{(a) The best SP-multiple-alignment created by the SP Computer Model with a store of Old SP-patterns like those in rows 1 to 8 (representing grammatical structures, including words) and a New SP-pattern (representing a sentence to be parsed) shown in row 0. Adapted with permission from Figure 1 in \protect\cite{wolff_sp_intelligent_database}.}
\label{parsing_figure}
\end{figure*}

\begin{figure*}[!htbp]
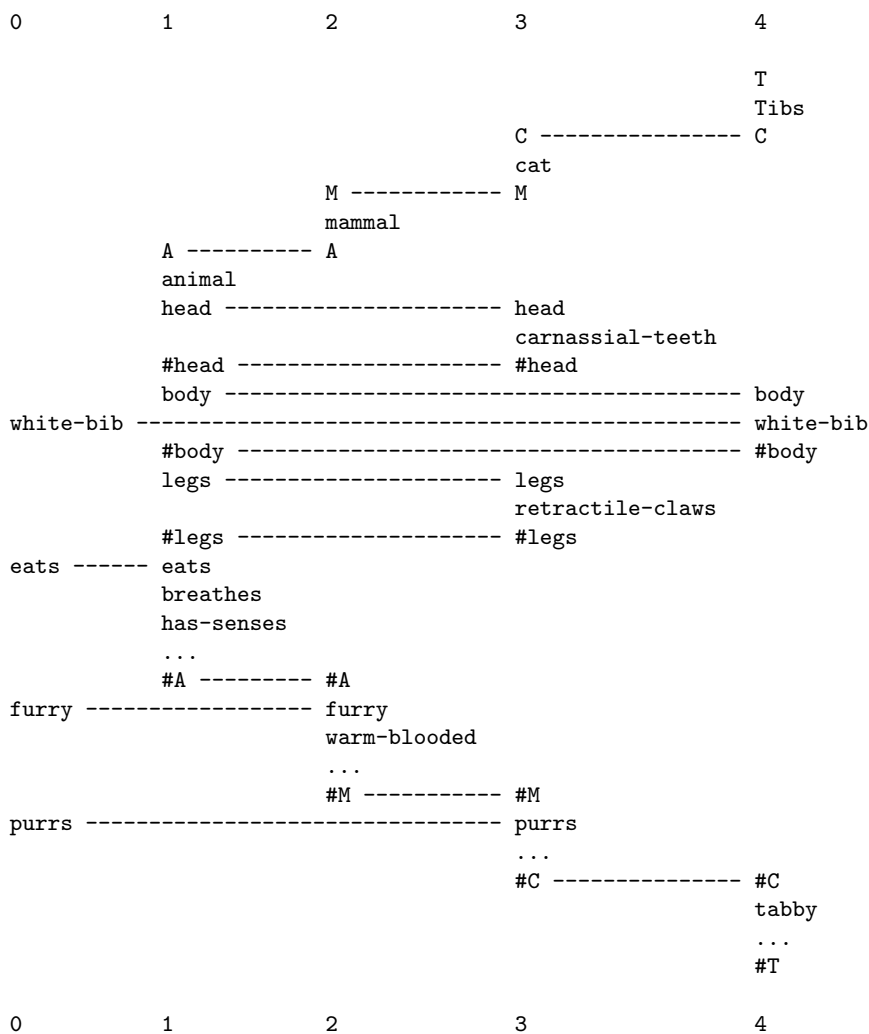

\fontsize{09.00pt}{10.80pt}
\centering
{\bf
\begin{BVerbatim}
0           1            2              3                  4

                                                           T
                                                           Tibs
                                        C ---------------- C
                                        cat
                         M ------------ M
                         mammal
            A ---------- A
            animal
            head ---------------------- head
                                        carnassial-teeth
            #head --------------------- #head
            body ----------------------------------------- body
white-bib ------------------------------------------------ white-bib
            #body ---------------------------------------- #body
            legs ---------------------- legs
                                        retractile-claws
            #legs --------------------- #legs
eats ------ eats
            breathes
            has-senses
            ...
            #A --------- #A
furry ------------------ furry
                         warm-blooded
                         ...
                         #M ----------- #M
purrs --------------------------------- purrs
                                        ...
                                        #C --------------- #C
                                                           tabby
                                                           ...
                                                           #T

0           1            2              3                  4
\end{BVerbatim}
}
\caption{The best SP-multiple-alignment found by the SP model, with the New SP-pattern `\texttt{white-bib eats furry purrs}' shown in column 1, and a set of Old SP-patterns representing different categories of animal and their attributes shown in columns 1 to 4. Reproduced with permission from Figure 15 in \protect\cite{sp_extended_overview}.}
\label{class_hierarchy_figure}
\end{figure*}

Examples of SP-multiple-alignments created by the SP Computer Model are shown in Figures \ref{parsing_figure} and \ref{class_hierarchy_figure}.

The first of these shows how the parsing of a sentence may be modelled within the SP-multiple-alignment framework. In the figure, row 0 shows a New SP-pattern representing the sentence to be parsed while rows 1 to 8 show Old SP-patterns, one per row, representing grammatical structures including words. Row 8 shows the grammatical dependency between the plural subject of the sentence (marked with `\texttt{Np}') and the plural main verb (marked with `\texttt{Vp}').

By contrast with the SP-multiple-alignment shown in Figure \ref{parsing_figure}, the SP-multiple-alignment in Figure \ref{class_hierarchy_figure} shows SP-patterns in columns, one SP-pattern per column, with alignments between SP-symbols shown in rows. The two ways of representing SP-multiple-alignments are entirely equivalent. The choice normally depends on what fits best on the page. Column 0 in Figure \ref{class_hierarchy_figure} shows a New SP-pattern representing something to be recognised while columns 1 to 4 show Old SP-patterns, one per row, representing categories of entity.

By convention, the New SP-pattern is always shown in column (or row) 0 and Old SP-patterns are shown in other columns (or rows). In some applications there is more than 1 New SP-pattern in column (row) 0.

\subsection{The SP71 Computer Model}\label{sp_computer_model_appendix}

The latest version of the SP Computer Model is SP71. Instructions for obtaining the source code are given in \ref{source_code_for_SP71_appendix}.

There is no comprehensive description of SP71, but SP70, a very similar and slightly earlier version of the program is described quite fully in \cite{wolff_2006}. More detailed citations are given in the three subsections that follow.

At the level of detail that will be considered, the description of SP70 in \cite{wolff_2006} is an accurate description of SP71. In what follows, it will be regarded as a description of SP71.

\subsubsection{Finding Good Full and Partial Matches Between SP-Patterns in SP71}\label{full_and_partial_matches_appendix}

In the foundations of the SP Computer Model is a process for finding good full and partial matches between two SP-patterns (sequences of symbols) that works well even when there are substantial differences between the two SP-patterns. This is described quite fully in \cite[Appendix A]{wolff_2006}.

This technique is similar in its effects to standard `dynamic programming' methods for comparing sequences \cite{wagner_fischer_1974,sankoff_kruskall_1983} but with these main advantages \cite[Section 3.10.3.1]{wolff_2006}:

\begin{itemize}

    \item It can match arbitrarily long sequences without excessive demands on memory.

    \item For any two sequences, it can find a set of alternative matches (each with a measure of how good it is) instead of a single `best' match.

    \item The `depth' or thoroughness of the searching can be controlled by parameters.

\end{itemize}

\subsubsection{Building SP-Multiple-Alignments in SP71}\label{building_multiple_alignments_appendix}

The way in which SP-multiple-alignments are built in SP71 is described in \cite[Sections 3.9 and 3.10]{wolff_2006}. Slightly confusingly, the main subject of these sections is SP61, but this is merely a part of the SP Computer Model that is concerned with the building of SP-multiple-alignments.

As with multiple sequence alignments in bioinformatics, and as with the process of finding good full and partial matches between two SP-patterns (Section \ref{full_and_partial_matches_appendix}), heuristic techniques are used in the building of SP-multiple-alignments in the SP System: searching for good SP-multiple-alignments in stages with a weeding out of the lower-scoring SP-multiple-alignments at the end of each stage.

\subsubsection{Unsupervised Learning in SP71}\label{unsupervised_learning_appendix}

Unsupervised learning in SP71 is described in \cite[Sections 3.9 and 9.2]{wolff_2006}, with relevant pseudocode in \cite[Figures 9.1 and 9.6]{wolff_2006}. The aim here is, for a given set of New SP-patterns, to create one or two {\em grammars}---meaning collections of Old SP-patterns---that are effective at encoding the given set of New SP-patterns in an economical manner.

The building of SP-multiple-alignments is an integral part of unsupervised learning in the SP System. It provides a means of creating Old SP-patterns via the splitting or amalgamation of pre-existing Old SP-patterns, and via the direct incorporation of New SP-patterns. And it provides a means of evaluating candidate grammars in terms of their effectiveness at encoding the given set of New SP-patterns in an economical manner.

As with the building of SP-multiple-alignments, and the process of finding good full and partial matches between SP-patterns, the creation of good grammars requires heuristic search through the space of alternative grammars: creating grammars in stages and discarding low-scoring grammars at the end of each stage.

\subsubsection{Varying the Thoroughness of Heuristic Search}\label{varying_the_thoroughness_of_search_appendix}

A useful feature of SP71 is that, with all three of the main components (\ref{full_and_partial_matches_appendix}, \ref{building_multiple_alignments_appendix}, and \ref{unsupervised_learning_appendix}), the thoroughness of the searches may be varied by varying the amount of memory space that is available for storing intermediate results. In effect, this controls the amount of backtracking that can be done and thereby controls the chances of escaping from local peaks in the search space.

\subsection{SP-Neural}\label{sp-neural_appendix}

Abstract concepts in the SP Theory map quite neatly into groupings of neurons and their interconnections in a version of the theory called {\em SP-Neural} \cite{spneural_2016}: SP-symbols are realised as {\em neural symbols} (single neurons or small clusters of neurons) and SP-patterns are realised as {\em pattern assemblies}. Although pattern assemblies in SP-Neural are quite similar to {\em cell assemblies} as described by Donald Hebb \cite{hebb_1949}, unsupervised learning in the SP System (including what is envisaged in the development of SP-Neural) is quite different from `Hebbian' learning \cite[p.~62]{hebb_1949} or versions of Hebbian learning that are popular in `deep learning in artificial neural networks' \cite{schmidhuber_2015}. The key differences are described in \cite[Section 10.5]{spneural_2016}.

\subsection{Distinctive Features and Advantages of the SP System}\label{sp_distinctive_features_advantages_appendix}

Distinctive features and advantages of the SP System are described in \cite{sp_alternatives}. In particular, Section V of that paper describes 13 problems with deep learning in artificial neural networks and how, with the SP System, those problems may be overcome.

The SP System also provides a comprehensive solution to a fourteenth problem with deep learning---`catastrophic forgetting'---meaning the way in which new learning in a deep learning system wipes out old memories.\footnote{A solution has been proposed in \cite{kirkpatrick_2017} but it appears to be partial, and it is unlikely to be satisfactory in the long run.}

There is further discussion in \cite{spdlsol_2018} of weaknesses in deep learning and how, in the SP System, those weaknesses may be overcome.

Key strengths of the SP System, which owe much to the central role in the system of the SP-multiple-alignment concept, are in three inter-related features of the system:

\begin{itemize}

    \item {\em Versatility in aspects of intelligence}. The SP System has strengths in several aspects of intelligence, including unsupervised learning, pattern recognition, planning, problem solving, and several kinds of reasoning, and there are reasons to think that SP-multiple-alignment may prove to be the key to general, human-like artificial intelligence. (\cite[Appendix B.2]{sp_software_engineering}, \cite[Section 4]{sp_intro_2018}).

    \item {\em Versatility in the representation of knowledge}. The SP System has been shown to be a versatile framework for the representation of diverse kinds of knowledge, and there are reasons to think that it may in principle accommodate any kind of knowledge (\cite[Appendix B.1]{sp_software_engineering}, \cite[Section 5]{sp_intro_2018}).

    \item {\em Seamless integration of diverse kinds of knowledge and diverse aspects of intelligence, in any combination}. Since the SP System's strengths in the representation of diverse kinds of knowledge and in diverse aspects of intelligence all flow from a single coherent source---the SP-multiple-alignment framework---there is clear potential for their seamless integration in any combination (as defined at the end of Section \ref{background_section}) \cite[Appendix B.3]{sp_software_engineering}. As was mentioned in Section \ref{background_section}, that kind of seamless integration appears to be essential in any system that aspires to human-like fluidity, versatility, and adaptability in intelligence.

\end{itemize}

\subsection{Potential Benefits and Applications of the SP System}\label{potential_benefits_applications_appendix}

As noted in Section \ref{overview_of_research_strategy_section}, the SP System has yielded several potential benefits and applications, described in a book and in several peer-reviewed papers:

\begin{itemize}

    \item The book, {\em Unifying Computing and Cognition} \cite{wolff_2006} describes strengths of the SP System in several areas, mainly in AI, with corresponding potential in applications that require human-like intelligence.

    \item The paper `The SP Theory of Intelligence: benefits and applications' \cite{sp_benefits_apps} describes several potential benefits and applications of the SP System, including potential for: an overall simplification of computing systems, including software; applications in the processing of natural language; benefits in software engineering; benefits in bioinformatics; and several more.

    \item The paper `Big data and the SP Theory of Intelligence' \cite{sp_big_data} describes how the SP System may help solve nine significant problems with big data.

    \item The paper `Autonomous robots and the SP Theory of Intelligence' \cite{sp_autonomous_robots} describes how the SP System may help in the development of human-like intelligence in autonomous robots.

    \item The paper `Application of the SP Theory of Intelligence to the understanding of natural vision and the development of computer vision' \cite{sp_vision} describes, as its title suggests, how the SP System may help in the understanding of human vision and in the development of computer vision.

    \item The paper `Towards an intelligent database system founded on the SP Theory of computing and cognition' \cite{wolff_sp_intelligent_database} describes how the SP System may function as a database system that, in a relatively streamlined manner, can accommodate a wider variety of kinds of knowledge than a conventional DBMS, with potential for diverse aspects of intelligence.

    \item The paper `Medical diagnosis as pattern recognition in a framework of information compression by multiple alignment, unification and search' \cite{wolff_medical_diagnosis} describes how the SP System may serve as a repository for medical knowledge and as a means of assisting medical practitioners in medical diagnosis.

    \item The paper `Commonsense reasoning, commonsense knowledge, and the SP Theory of Intelligence' \cite{sp_csrk} describes how the SP System may help solve several of the problems in modelling commonsense reasoning and commonsense knowledge, described by Ernest Davis and Gary Marcus in \cite{davis_marcus_2015}.

    \item The draft report `Software engineering and the SP Theory of Intelligence' \cite{sp_software_engineering} describes how the SP System may be applied in software engineering, and the potential benefits.

\end{itemize}

\section{Source Code and Windows Executable Code for the SP71 Computer Model}\label{source_code_for_SP71_appendix}

The latest version of the SP Computer Model is SP71. At the time of writing, the source code, written in C++ with many comments, may be obtained via instructions under the headings `Source code' and `Archiving', near the bottom of \href{http://bit.ly/1mSs5XT}{bit.ly/1mSs5XT}. Likewise for the Windows executable code. More specifically, these things may be obtained:

\begin{itemize}

    \item In the file SP71.zip (\href{http://bit.ly/1OEVPDw}{bit.ly/1OEVPDw}).

    \item The source code is in `Ancillary files' with `The SP Theory of Intelligence: an overview' under \href{http://www.arxiv.org/abs/1306.3888}{www.arxiv.org/abs/1306.3888}. Apart from SP71.exe, all the files should be treated as plain text files to be opened with Wordpad, WinEdt, or the like.

    \item From a digital archive with: the National Museum of Computing, Block H, Bletchley Park, Milton Keynes, MK3 6EB, UK; {\em Phone:} +44 (0)1908 374708; {\em Email:} operations@tnmoc.org.

    \item From a digital archive with: the Library and Archives Service, Bangor University, Bangor, Gwynedd, LL57 2DG, UK; {\em Phone:} +44 (0) 1248 382981; {\em Email:} library@bangor.ac.uk.

\end{itemize}

\bibliographystyle{compj}

\end{document}